%% file: main.tex
\documentclass[a4paper,11pt]{article}

\usepackage{tikz-qtree}
\usepackage{tikz-qtree-compat}
\usepackage{tikz}
\usetikzlibrary{shapes,arrows,shadows,calc,decorations.pathreplacing,decorations.markings}
\usepackage{amsmath,amssymb,bm,times}
\usepackage{graphicx}  %
\usepackage{url}       %
\usepackage{times}
\usepackage{natbib}
\usepackage[margin=25mm]{geometry}

\usepackage{wrapfig}
\usepackage{epsfig}
\usepackage{subcaption}
\usepackage{calc}
\usepackage{amssymb}
\usepackage{amstext}
\usepackage{amsmath}
\usepackage{amsthm}
\usepackage{multicol}
\usepackage{pslatex}
\usepackage{enumitem}
\usepackage{multirow}
\usepackage{mathabx}
\usepackage[multiple,bottom]{footmisc} %
\usepackage{xcolor}
\usepackage[T1]{fontenc}
\usepackage{inconsolata}
\usepackage{hyperref}

\usepackage{ifthen,xcolor}
\newlength{\tabcont}
\newcommand{\tab}[1]{%
\settowidth{\tabcont}{#1}%
\ifthenelse{\lengthtest{\tabcont < .25\linewidth}}%
{\makebox[.25\linewidth][l]{#1}\ignorespaces}%
{\makebox[.5\linewidth][l]{#1}\ignorespaces}%
}%

\usepackage{framed}

\newcommand{\lm}{\ensuremath{\lambda}}

\newcommand{\rar}{\ensuremath{\rightarrow}}

\newcommand{\mono}[1]{\texttt{#1}}
\newcommand{\ulf}[1]{\mono{#1}} %
\usepackage{multido}
\newcommand{\Repeat}{\multido{\i=1+1}}
\newcommand{\numCharTab}[1]{\Repeat{#1}{\phantom{a}}}

\newcommand{\dom}{\ensuremath{\mathcal{D}}}
\newcommand{\sit}{\ensuremath{\mathcal{S}}}
\newcommand{\tru}{\ensuremath{\text{\textbf{2}}}}

\newcommand{\prd}{\ensuremath{\mathcal{N}}} %
\newcommand{\wld}{\ensuremath{\mathcal{W}}}
\newcommand{\prp}{\ensuremath{\mathcal{P}}}
\newcommand{\knd}{\ensuremath{\mathcal{K}}}

\newcommand{\model}{\ensuremath{\mathcal{M}}}
\newcommand{\varas}{\ensuremath{\mathcal{U}}}

\newcommand{\act}{\ensuremath{\mathcal{A}}}

\newcommand{\atemptype}{\ensuremath{\dom \rar \tru}}
\newcommand{\temptype}{\ensuremath{\dom \rar \sit \rar \tru}}

\newcommand{\ulfwebsite}{\url{http://cs.rochester.edu/u/gkim21/ulf/}}

\title{A Type-coherent, Expressive Representation as an Initial Step to Language Understanding}
\date{}

\author{Gene Louis Kim and Lenhart Schubert\\
    Department of Computer Science, University of Rochester\\
\texttt{\{gkim21,schubert\}@cs.rochester.edu}
}

\begin{document}

\maketitle
\thispagestyle{empty}
\pagestyle{empty}

\begin{abstract}
A growing interest in tasks involving language understanding by the NLP community has led to the need for effective semantic parsing and inference. Modern NLP systems use semantic representations that do not quite fulfill the nuanced needs for language understanding: adequately modeling language semantics, enabling general inferences, and being accurately recoverable.  This document describes underspecified logical forms~(ULF) for Episodic Logic~(EL), which is an initial form for a semantic representation that balances these needs.  ULFs fully resolve the semantic type structure while leaving issues such as quantifier scope, word sense, and anaphora unresolved; they provide a starting point for further resolution into EL, and enable certain structural inferences without further resolution. This document also presents preliminary results of creating a hand-annotated corpus of ULFs for the purpose of training a precise ULF parser, showing a three-person pairwise interannotator agreement of 0.88 on confident annotations. We hypothesize that a divide-and-conquer approach to semantic parsing starting with derivation of ULFs will lead to semantic analyses that do justice to subtle aspects of linguistic meaning, and will enable construction of more accurate semantic parsers.
\end{abstract}

\section{Introduction}
\label{sec:introduction}

\begin{figure}[t]
\vspace*{-0.05in}
\centering
\input{semantic-interpretation.tex}
\vspace*{-0.2in}
\setlength{\belowcaptionskip}{-15pt}
{\small \caption{\label{fig:sem-pipeline}{\small The semantic interpretation process, with the ULF step in the fore.  Structurally dependent steps in the interpretation process are connected by solid black arrows and structurally independent information flow is represented with dashed blue arrows. The components that changed from the previous structural step are highlighted in yellow. Backward information arrows indicate that arriving at the optimal choice at a particular step may depend on ``later'' -- or structurally dependent -- steps.}}}
\normalsize
\end{figure}
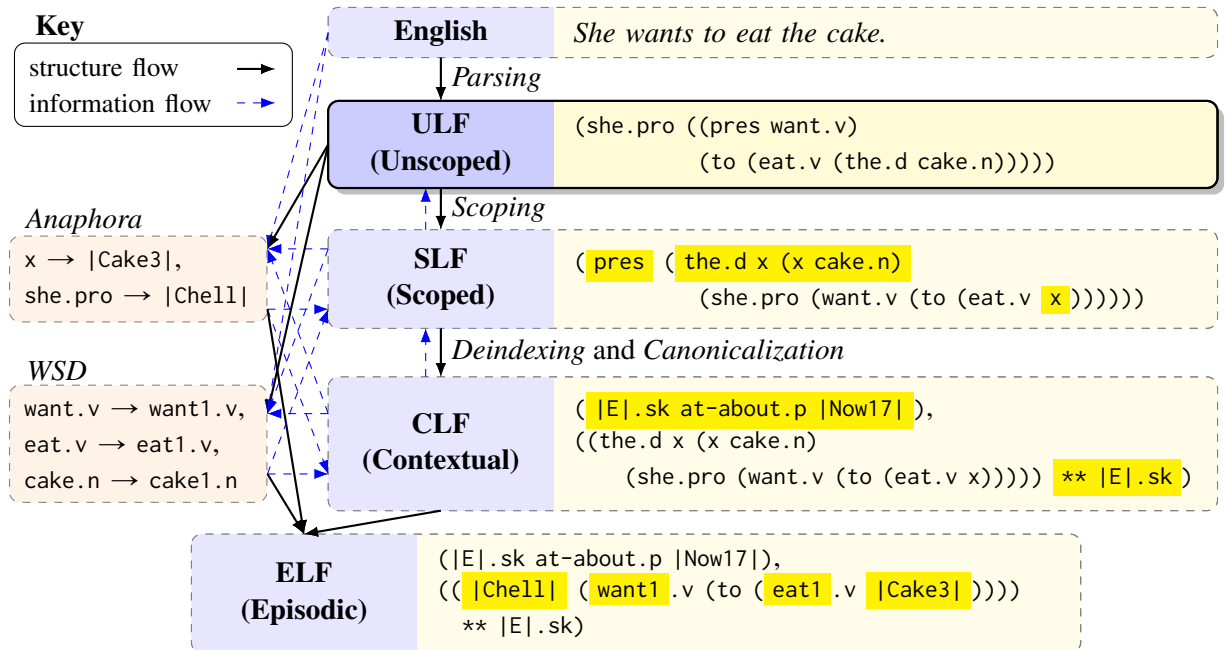

\noindent
Episodic Logic~(EL) is a semantic representation extending FOL, designed to closely match the expressivity and surface form of natural language and to enable deductive inference, uncertain inference, and NLog-like inference~\citep{morbini2009LFCR,schubert2000book,schubert2014SP}. 
\cite{kim2016*SEM}
developed a system that transforms annotated WordNet glosses into EL axioms which were competitive with state-of-the-art lexical inference systems while achieving greater expressivity.  While EL is representationally appropriate for language understanding, the current EL parser is
too unreliable for general text: The phrase structures produced by the underlying Treebank
parser leave many ambiguities in the semantic type structure, which are disambiguated 
incorrectly by the hand-coded compositional rules; moreover, errors in the phrase structures 
can further disrupt the resulting logical forms~(LFs).
\cite{kim2016*SEM} discuss the limitations of the existing parser 
as a starting point for logically interpreting
glosses of WordNet verb entries.  In order to build a better EL parser, it seems natural to take advantage of recent advances in corpus-based parsing techniques. 

This document describes a type-coherent initial LF, or \textit{unscoped logical forms}~(ULF), for EL which captures the predicate-argument structure in the EL semantic types and is the first critical step in fully-resolved semantic interpretation of sentences. 
Montague's profoundly influential work~\citep{montague1973} demonstrates that systematic assignments of appropriate semantic types to words and phrases allows us to view language as akin to formal logic, with meanings determined compositionally from syntactic structures.  This view of language directly supports inferences, at least to the extent that we can resolve -- or are prepared to tolerate -- ambiguity, context-dependence, and indexicality, towards which semantic types are agnostic. 
ULF takes a minimal step across the syntax-semantics interface by doing exactly this -- selecting the semantic types of words within EL. Thus ULFs are amenable to corpus-construction and statistical parsing using techniques similar to those used for syntax, and they enable generation of context-dependent structural inferences.  The nature of these inferences is discussed in more detail in Section~\ref{sec:inf-with-ulfs}.

Our working hypothesis in designing ULF is that a divide-and-conquer approach starting with preliminary surface-like LFs is a practical way to generate fully resolved interpretations of natural language in EL.  Figure~\ref{fig:sem-pipeline} shows a diagram of our divide-and-conquer approach, which elaborated upon in Section~\ref{sec:ulf-in-interp}.
We also outline a framework for quickly and reliably collecting ULF annotations for a corpus in a multi-pronged approach.  Our evaluation of the annotation framework shows that we achieve annotation speeds and agreement comparable to those for the \textit{abstract meaning representation}~(AMR) project, which has successfully built a large enough corpus to drive research into corpus-based parsing~\citep{banarescu2013LAW}. 
Further resources relating to this project, including a more in-depth description of ULFs, the annotation guidelines, and related code are available from the project website~\ulfwebsite.

\section{Episodic Logic}
\label{sec:el}

EL is a semantic representation that extends FOL to more closely match the expressivity of natural languages. It echoes both the surface form of language, and more crucially, the semantic types that are found in all languages. Some semantic theorists view the fact that noun phrases denoting both concrete and abstract entities can appear as predicate arguments (\mbox{\textit{Aristotle}}, \mbox{\textit{everyone}}, \textit{the fact that there is water on Mars}) as grounds for treating all noun phrases as being of higher types (e.g., second-order predicates). EL instead uses a small number of reification operators to map predicate and sentence intensions to individuals. As a result, quantification remains first-order~(but allows quantified phrases such as \textit{most people who smoke}, or \textit{hardly any errors}). Another distinctive feature of EL is that it treats the relation between sentences and episodes (including events, situations, and processes) as a \textit{characterizing} relation, written `**'. This coincides with the Davidsonian treatment of events as extra variables of predicates~\citep{davidson1967LDA} when we restrict ourselves to positive, atomic predications. However, `**' also allows for logically complex characterizations of episodes, such as \textit{not eating anything all day}, or \textit{each superpower menacing the other with its nuclear arsenal}~\citep{schubert2000book2}.

EL defines a hierarchical ontology over the domain of individuals, \dom.  \dom\ includes simple individuals, e.g. \textit{John}, possible situations, \sit,  possible worlds, \wld $\subset$\sit, various numerical types, propositions, \prp, and kinds, \knd, as well as others that are not important for the purposes of this document. 
A complete description of the ontology is provided by~\cite{schubert2000book}. The types of some predicates are further restricted by these categories.  For example, the predicate \ulf{claim.v} -- as in \textit{``I claim that grass is red.''} -- has the type $\prp \rar (\dom \rar (\sit \rar \tru))$, since its first argument is a proposition and the second argument is a simple individual~(in the semantics of EL the agent argument is supplied last, though it precedes the predicate in the surface syntax).   

The semantic types in EL are defined by recursive functions over individuals, \dom, and truth values, $\{0,1\}$, written as \tru.  Semantic values of predicates applied to their surface arguments can yield a value in \tru\ at a given (possible) situation, or be \textit{undefined} there (indicating irrelevance of the predication in the given situation).  Most predicates in EL are of type $\dom^n \rar (\sit \rar \tru)$ (where $\dom^2 \rar \tru$ abbreviates $\dom \rar (\dom \rar \tru)$, $\dom^3 \rar \tru$ abbreviates $\dom \rar (\dom \rar (\dom \rar \tru))$, and so on).  That is, they are first-order intensional predicates.\footnote{Some predicates allow for a monadic predicate complement such as \textit{look} in \textit{``They look happy''}.} Monadic predicates play a particularly important role in EL as well as ULF, and we will abbreviate their type $\dom \rar (\sit \rar \tru)$ as \prd. In EL syntax, square brackets indicate infixed operators~(i.e.$\; [\tau_n \; \pi \; \tau_1 \; ... \; \tau_{n-1}]$ where $\pi$ is the operator) and parentheses indicate prefixed operators~(i.e.$ \; (\pi \; \tau_1 \; ... \; \tau_{n})$ where $\pi$ is the operator).  
Predicative formulas such as \ulf{[|Aristotle| famous.a]} or \ulf{[|Romeo| love.v |Juliet|]} are regarded as temporal and must be evaluated with respect to a situation via an episode-relating operator~(e.g. `**') to supply the episode and thus produce an atemporal formula.

There are also a limited number of type-shifting operators in EL to map between some of these types. The kind operator, `\ulf{k}', shifts a monadic predicate into a kind, $(\dom \rar (\sit \rar \tru)) \rar \knd$, and the operator , `\ulf{that}', forms propositions from sentence intensions, $(\sit \rar \tru) \rar \prp$. \textit{``that grass is red''}, a segment of an earlier example, is formulated as \ulf{(that [(k grass.n) red.a])} in EL, uses both of these operators.

\section{Unscoped logical form}
\label{sec:ulf-intro}

ULFs are type-coherent initial LFs which provide a stepping stone to capturing full sentential EL meanings. They enable interesting classes of structural inferences that are of broader scope than those enabled by Natural Logic~(NLog)~\citep{sanchez1995LA}, and unlike NLog inferences do not depend on prior knowledge of the propositions to be confirmed or refuted.  ULF captures the full predicate argument structure of EL while leaving word sense, scope, and anaphora unresolved.  Therefore, ULFs can be analyzed using the formal EL type system while taking the scopal ambiguities into account. There is not enough space here to exhaustively discuss how ULF handles various phenomena, so the discussion will be restricted to the broad framework of ULF and the most crucial aspects of the semantics. Please refer to~\ulfwebsite\ for complete information on ULF.

\subsection{ULF Syntax}
\label{sec:ulf-syntax}

All atoms in ULF, with the exception of certain logical functions and syntactic macros, are marked with an atomic syntactic type.  The atomic syntactic types are written with suffixed tags: \ulf{.v,.n,.a,.p,} \ulf{.pro,.d,.aux-v,.aux-s,.adv-a,.adv-e,.adv-s,.adv-f,.cc,.ps,.pq,.mod-n,} or \ulf{.mod-a}, except for names, which use wrapped bars, e.g. \ulf{|John|}.  These are intended to echo the part-of-speech origins of the constituents, such as \textit{verb, noun, adjective, preposition, pronoun, determiner,} etc., respectively; some of them contain further specifications as relevant to their entailments, e.g., \ulf{.adv-e} for locative or temporal adverbs (implying properties of events).  The distinctions among predicates of sorts \ulf{.v,.n,.a,.p}, corresponding to English parts of speech, are often suppressed in other LFs for language, but are semantically important. For example, \textit{``Bob danced"} can refer to a brief episode while \textit{``Jill was a dancer"} generally cannot (and may suggest Jill is no longer alive); this is related to the fact that verbal predicates are typically ``stage-level" (episodic) while nominal predicates are generally ``individual-level" (enduring). 
Whereas in EL the bracket type specifies whether prefix or infix notation is being used, in ULF this distinction is inferred from the semantic types of the constituents and only parentheses are used. Atoms that are implicit in the sentence or elided and thus supplied by the annotator are wrapped in curly brackets, such as \ulf{\{ref\}.pro} in example~\ref{itm:ulf-dial-example} of Figure~\ref{fig:ulf-examples}.

\begin{figure}[t]
\begin{framed}
  \vspace{-5pt}
{\small
  \begin{enumerate}[label=(\arabic*),itemsep=-1pt]
  \item \textit{Could you dial for me?} 
  \label{itm:ulf-dial-example}\\
  \ulf{(((pres could.aux-v) you.pro (dial.v \{ref1\}.pro
                                     (adv-a (for.p me.pro)))) ?)}

  \item \textit{If I were you I would be able to succeed.} 
  \label{itm:ulf-cf-example}\\
  \ulf{((if.ps (i.pro ((cf were.v) (= you.pro))))\\
  \numCharTab{1}(i.pro ((cf will.aux-s) (be.v (able.a (to succeed.v))))))}

  \item \textit{Flowers are weak creatures} 
  \label{itm:ulf-flowers-example}
  
  \vspace*{-0.08in}
  
  \ulf{((k (plur flower.n)) ((pres be.v) 
                                (weak.a (plur creature.n))))}

  \vspace*{0.02in}

  \item \textit{Very few people still debate the fact that the earth is heating up} 
  \label{itm:ulf-earth-example}
\vspace*{-.13in}
\small
\begin{verbatim}
(((fquan (very.mod-a few.a)) (plur person.n)) 
          (still.adv-s (debate.v
             (the.d (n+preds fact.n (= (that ((the.d |Earth|.n) 
                                              ((pres prog) heat_up.v))))))))
\end{verbatim}
\vspace*{-0.33in}
  \end{enumerate}
}
\vspace{5pt}
\end{framed}
\vspace{-18pt}
  \setlength{\belowcaptionskip}{-15pt} %
{\small 
  \caption{\label{fig:ulf-examples}{\small Example sentences with corresponding raw ULF annotations.  Examples \ref{itm:ulf-dial-example} and \ref{itm:ulf-cf-example} are from the Tatoeba database, \ref{itm:ulf-flowers-example} is from  \textit{The Little Prince}, and \ref{itm:ulf-earth-example} is from the Web.}}}
  \normalsize
 \end{figure}

For practical purposes we distinguish \textit{raw ULF} from \textit{postprocessed ULF}. In raw ULF we allow certain argument-taking constituents to be dislocated from their ``proper" place, so as to adhere more closely to linguistic surface structure and thereby facilitate annotation. For example, 
sentence-level operators (of type \ulf{adv-s}) appearing mid-sentence may be left ``floating''~(e.g., \ulf{(|Alice| certainly.adv-s ((pres know.v) |Bob|))}), since they can be automatically lifted to the sentence-level; and verb-level adverbs (of type \ulf{adv-a}) can be interleaved with arguments (e.g., \ulf{((past speak.v) sternly.adv-a (to.p-arg |Bob|))}), even though semantically they  operate on the whole verb phrase.  \cite{kim2017SemBEaR} presented this method of dislocated annotation for sentence-level operators. 
In postprocessed ULF, we can understand all atoms and subexpressions of well-formed formulas~(wffs) as being one of the following ULF constituent types (modulo some following remarks):

\textit{entity, predicate, determiner, monadic predicate modifier, sentence, sentence modifier,\\
\indent connective, lambda abstract, or one of a limited number of type-shifting operators},

\noindent
where the predicates and operators that act on predicates are subcategorized by whether the predicate is derived from a noun, verb, adjective, or preposition.  These constituent types uniquely map to particular semantic types, i.e. are aliases for the formal types.  Clausal constituents are combined according to their bracketing and semantic types.  

A qualification of the above general claim is that unscoped tense operators, determiners, and coordinators remain in their surface position even in postprocessed ULF.
For example, in \ulf{(|Bob| ((pres own.v) (a.d dog.n)))}, \ulf{pres} is actually an unscoped sentence-level operator (which, in conversion to EL, is deindexed to yield a characterization of an episode by the sentence, and a temporal predication about that episode). We also retain coordinated expressions such as \ulf{((in.p |Rome|) and.cc happy.a)}, where this will ultimately lead to a sentential conjunction in EL. Similarly, \ulf{(a.d dog.n)} is kept in argument position as if it were of semantic type \dom\ (thus, as if the determiner were of semantic type $\prd \rar \dom$).\footnote{The actual semantic type of determiners in EL, after lambda-abstraction of the restrictor and matrix formula, is $\prd \rar (\prd \rar (\sit \rar \tru))$.  See Appendix~\ref{app:quantifier-semantics} for full details.} Such unscoped constituents do not disrupt type coherence, because the possible conversions to type-coherent EL are well-defined. 

 Finally, both raw ULFs and postprocessed ULFs can contain macros. For example, the macro operator \ulf{n+preds} is used for postmodified nominal predicates such as \ulf{(n+preds dog.n (on.p (a.d leash.n)))} -- see also example~\ref{itm:ulf-earth-example} in Figure~\ref{fig:ulf-examples}; this avoids immediate introduction of a \lm-abstracted conjunction of predicates, simplifying the annotation task. Appendix~\ref{app:macro} discusses macros further, including their formal definitions.
Section~\ref{sec:ulf-details} will ground the high-level discussions in this and the following section with a concrete discussion of modifiers.

\subsection{ULF Type Structure}
\label{sec:type-structure}

The type-shifting operators mentioned in the previous section are crucial for type coherence in ULFs. In example~\ref{itm:ulf-dial-example} the phrase \textit{``for me''} is coded as \ulf{(adv-a (for.p me.pro))}, rather than simply \ulf{(for.p me.pro)}  because it is functioning as a \textit{predicate modifier}, semantically operating on the verbal predicate \ulf{(dial.v \{ref1\}.pro)} (\textit{dial a certain thing}).   Let $\prd_{ADJ}$, $\prd_{N}$, and $\prd_{V}$ be the sortal refinements of the monadic predicate type \prd\ corresponding to adjectives, nouns, and verbs, respectively.  \ulf{(adv-a (for.p me.pro))} has type $\prd_{V} \rar \prd_{V}$.  Without the \ulf{adv-a} operator the prepositional phrase is just a 1-place predicate. Its use as a predicate is apparent in contexts like \textit{``This puppy is for me"}. Note that semantically the 1-place predicate \ulf{(for.p me.pro)} is formed by applying the 2-place predicate \ulf{for.p} to the (individual-denoting) term \ulf{me.pro}. 
If we apply \ulf{(for.p me.pro)} to another argument, such as \ulf{|Snoopy|} (the name of a puppy), we obtain a sentence intension.\footnote{\ulf{(for.p me.pro)} has type $\dom \rar (\sit \rar \tru)$ and \ulf{|Snoopy|} has type $\dom$, so \ulf{(|Snoopy| (for.p me.pro))} has a type that resolves to $\sit \rar \tru$~(i.e. a sentence intension).} So semantically, \ulf{adv-a} is a \textit{type-shifting operator} of type $\prd \rar (\prd_{V} \rar \prd_{V})$. 

This brings up the issue of \textit{intensionality}, which is preserved in ULF. Example~\ref{itm:ulf-cf-example} is a counterfactual conditional, and the consequent clause \textit{``I would be able to succeed"} is not evaluated in the actual world, but in a possible world where the (patently false) antecedent is imagined to be true. ULF captures this with the `\ulf{cf}' operator in place of the tense and the EL formulas derived from it are evaluated with respect to \textit{possible situations (episodes)}, whose maxima are possible worlds.   The type of `\ulf{cf}' is $(\sit \rar \tru) \rar (\sit \rar \tru)$ after operator scoping to the sentence-level, but like tense operators is kept with the verb in raw ULF, essentially functioning as a predicate-level identity function, $(\lm X.X)$, there.

`\ulf{to}' in~\ref{itm:ulf-cf-example}, `\ulf{k}' in~\ref{itm:ulf-flowers-example}, and `\ulf{that}' in~\ref{itm:ulf-earth-example} are all operators that reify different semantic categories, shifting them to abstract individuals. `\ulf{to}' (synonym: \ulf{ka}) shifts a verbal predicate to a \textit{kind (type) of action or attribute}, $\prd_{V} \rar \knd_{A}$; `\ulf{k}' shifts a nominal predicate to a \textit{kind} of thing, $\prd_{N} \rar \knd$ (so the subject in example~\ref{itm:ulf-flowers-example} is the abstract kind, \textit{flowers}, whose instances consist of sets of flowers); and `\ulf{that}' produces a reified \textit{proposition}, $(\sit \rar \tru) \rar \prp$ (again an abstract individual) from a sentence meaning. Using these type shifts, EL and ULF are able to maintain a simple, classical view of predication, while allowing greater expressivity than the most widely employed LFs.%

\subsection{Role of ULF in Comprehensive Semantic Interpretation}
\label{sec:ulf-in-interp}

ULFs are underspecified, but their surface-like form and the type structure they encode make them well-suited to reducing underspecification by using well-established linguistic principles and exploiting the distributional properties of language.
Figure~\ref{fig:sem-pipeline} shows the interpretation process for EL formulas and the role of ULFs in providing the first step into it.  Due to the structural dependencies between the components in the interpretation process, the optimal choice at any given component depends on the overall coherence of the final interpretation; hence the backward arrows in the figure.  Word sense disambiguation~(WSD) and anaphora have no structural dependencies in the interpretation process so they are separated from and fully connected to the post-ULF components.  These resolutions are depicted in the last step in the figure.

\vspace*{0.3em}

\noindent
\textbf{WSD \& Anaphora}: 
While \ulf{(weak.a (plur creature.n))} in example~\ref{itm:ulf-flowers-example} does not specify which of the dozen WordNet senses of \textit{weak} or three senses of \textit{creature} is intended here, the type structure is perfectly clear: A predicate modifier is being applied to a nominal predicate. 
ULF also does not assume unique adicity of word-derived predicates such as \ulf{run.v}, since such predicates can have intransitive, simple transitive and other variants, but the
adicity of a predicate in ULF is always clear from its structural context -- we know that it has all its arguments in place
when an argument (the ``subject") is placed on its left, as in English. 

Linguistic constraints (e.g. \textit{binding constraints}) exist for coreference resolution. For example, in \textit{``John said that he was robbed"}, \textit{he} can refer to \textit{John}; but this is not possible in \textit{``He said that John was robbed"}, because in the latter, \textit{he} C-commands \textit{John}, i.e., in the phrase structure of the sentence, it is a sibling of an ancestor of \textit{John}. 
ULF preserves this structure, allowing use of such constraints.
While ULF constrains the word senses and coreferences through adicity and syntactic structure, WSD and anaphora resolution should not be applied to isolated sentences since word sense patterns and coreference chains often span multiple sentences.

\vspace*{0.3em}

\noindent
\textbf{Scoping:} %
Unscoped constituents~(determiners, tense operators, and coordinators) can generally ``float" to more than one possible position. 
Following a view of scope ambiguity developed by \cite{schubert1982CL} elaborated on by \cite{hurum1986AI}, these constituents always float to pre-sentential positions, and determiner phrases leave behind a variable
that is then bound at the sentential level. 
The accessible positions are constrained by linguistic restrictions,
such as \textit{scope island} constraints in subordinate clauses~\citep{ruys2010HPL}.
Beyond this, many factors influence preferred scoping possibilities, with surface form playing a prominent role~\citep{manshadi2013ACL}.  
The proximity of ULF to surface syntax enables the use of these constraints. 

\vspace*{0.3em}

\noindent
\textbf{Deindexing and Canonicalization:} %
Much of the past work relating to EL has been concerned with the principles of \textit{deindexing}%
~\citep{hwang1992thesis,hwang1994ICTL,schubert2000book}.  Deindexing corresponds to the introduction of event variables for explicitly characterizing the sentence it is linked to via the `\ulf{**}' operator~(this variable becomes \ulf{|E|.sk} in Figure~\ref{fig:sem-pipeline} after Skolemization).
Hwang and Schubert's approach to tense-aspect processing, constructing \textit{tense trees} for temporally relating event variables, is only possible if the LF being processed reflects the original clausal structure -- as ULF indeed does.
Canonicalization is the mapping of an LF into ``minimal'', distinct propositions, with top-level Skolemization. %
The CLF step in Figure~\ref{fig:sem-pipeline} contains two separate formulas as a result of this process.

\vspace*{0.3em}

\noindent
\textbf{Episodic Logical Forms (ELF):} When episodes have been made explicit 
and all anaphoric and word ambiguities are resolved the result is a set of \textit{episodic logical forms}.  These can be used in the \textsc{Epilog} inference engine for reasoning that combines linguistic semantic content with world knowledge.\footnote{\textsc{Epilog} is competitive against state-of-the-art FOL theorem provers~\citep{morbini2009LFCR}.} 
A variety of complex \textsc{Epilog} inferences are reported by~\cite{schubert2013LiLT}, 
and \cite{morbini2011chapter} give examples of self-aware metareasoning. 
\textsc{Epilog} also reasoned about snippets from the Little Red 
Riding Hood story, for example using knowledge about the world and goal-oriented
behavior to understand why the presence of nearby woodcutters prevented 
the wolf from attacking Little Red Riding Hood when he first saw her~\citep{hwang1992thesis,schubert2000book}.

\subsection{Inference with ULFs}
\label{sec:inf-with-ulfs}

An important insight of NLog research is that language can be used
directly for inference, requiring only phrase structure analysis 
and upward/downward entailment marking (polarity) of phrasal contexts. 
This means that NLog inferences are \textit{situated} inferences, i.e.,
their meaning is just as dependent on the utterance setting and discourse
state as the linguistic ``input" that drives them.  
This insight carries over to ULFs, and provides a separate justification
for computing ULFs, apart from their utility in the process of deriving
EL interpretations 
from language. The semantic type structure encoded by ULFs provides a more reliable and 
general basis for situated inference than mere phrase structure. 
Here, briefly, are some kinds of inferences we can expect ULFs to support with minimal additional knowledge due to their structural nature:
\begin{itemize}%
 \item \textit{NLog inferences based on generalizations/specializations}.
   For example, \textit{``Every NATO member sent troops to Afghanistan"}, 
   together with the knowledge that France is a NATO member and that 
   Afghanistan is a country entails that \textit{France sent troops to
   Afghanistan} and that \textit{France sent troops to a country}.
 \item \textit{Inferences based on implicatives}. For example, \textit{``She
   managed to quit smoking"} entails that \textit{She quit smoking} (and
   the negation of the premise leads to the opposite conclusion). Inferences of this sort have
   been demonstrated for headlines using ELFs by \cite{stratos2011KEOD}.  %
 \item \textit{Inferences based on attitudinal and communicative verbs}.
   For example, \textit{``John denounced Bill as a charlatan"}
   entails that \textit{John probably believes that Bill is a charlatan},
   that \textit{John asserted to his listeners (or readers) that Bill is 
   a charlatan}, and that \textit{John wanted his listeners (or readers)
   to believe that Bill is a charlatan}. 
   These inferences would be hard to capture within NLog, since they are partially probabilistic, require structural elaboration, and depend on constituent types.
 \item \textit{Inferences based on counterfactuals}. For example, \textit{``If
   I were rich, I would pay off your debt"} and \textit{``I wish I were rich"}
   both implicate that \textit{the speaker is not rich}. This depends 
   on recognition of the counterfactual form, which is distinguished
   in ULF.
 \item \textit{Inferences from questions and requests}. For example, \textit{
   ``When are you getting married?"} enables the inferences that the 
   addressee will get married (in the foreseeable future), that 
   the questioner wants to know the expected date of the event, and
   that the addressee probably knows the answer and will supply
   it. Similarly an apparent request such as \textit{``Could you close the 
   door?"} implies that the speaker wants the addressee to close the 
   door, and expects that he or she will do so. 
   
\end{itemize}

\section{Predicate and Sentence Modification in Depth}
\label{sec:ulf-details}

Here we ground the general description of ULF given so far with an in-depth discussion of how ULF handles modification.  This is done with the purpose of demonstrating how the core syntax of ULF, its syntactic looseness, and semantic types fit together in practice.
EL semantic types represent \textit{predicate modifiers} as functions from
\textit{monadic} intensional predicates to \textit{monadic} intensional predicates, i.e., 
$\prd \rar \prd$, which enables handling of intersective, subsective, and 
intensional modifiers such as in the examples

  \ulf{((mod-n wooden.a) shoe.n), ((mod-n ice.n) pick.n), (fake.mod-n ruby.n),}\\
  \indent \ulf{((mod-a worldly.a) wise.a), (very.mod-a fit.a), (slyly.adv-a grin.v)}.

\noindent
Modifier extensions \ulf{.mod-n}, and \ulf{.mod-a} respectively reflect the linguistic 
categories of noun-premodifying (attributive) adjectives and adjective-premodifying
adverbs; correspondingly, operators \ulf{mod-n}, and \ulf{mod-a} type-shift 
prenominal predicates to modifiers applicable to predicates of sorts \ulf{.n} and 
\ulf{.a} respectively. Modifier extension \ulf{.adv-a} reflects the linguistic 
category of VP adverbials, and operator \ulf{adv-a} creates such modifiers from 
predicates. Thus, \textit{``walk with Bob"} is represented in raw 
and postprocessed ULF respectively as

  \ulf{(walk.v (adv-a (with.p |Bob|)))} and \ulf{((adv-a (with.p |Bob|)) walk.v)}.
  
\noindent
Adverbial modifiers of the sort \ulf{.adv-a} intuitively modify actions, experiences,
or attributes, as distinct from events. Thus \textit{``He lifted the child 
\underline{easily}"} refers to an action that was easy
for the agent, 
rather than to an easy event.
Actions, experiences, and attributes in EL are individuals comprised of agent-episode
pairs, and this allows
modifiers of the sort \ulf{.adv-a} to express a constraint on both the agent 
and the episode it characterizes. 
As such, actions are not explicitly represented in ULF and derived during deindexing when event variables are introduced. 

A formula or nonatomic verbal predicate in ULF may contain sentential
modifiers of type $(\mathcal{S} \rar \tru) \rar (\mathcal{S} \rar \tru)$:
\ulf{.adv-s}, \ulf{.adv-e}, and \ulf{.adv-f}. Again there are type-shifting operators that create these sorts of modifiers 
from monadic predicates. Ones of the sort \ulf{.adv-s} are usually modal (and thus opaque), e.g., 

   \ulf{perhaps.adv-s}, \ulf{(adv-s (without.p (a.d doubt.n)))};
   
\noindent
However, negation is transparent in the usual sense -- the truth value of a negated sentence depends only of the truth value of the unnegated sentence. 
Modifiers of sort \ulf{.adv-e} are transparent, typically implying temporal or locative
constraints, e.g.,

   \ulf{today.adv-e}, \ulf{(adv-e (during.p (the.d drought.n)))}, \ulf{(adv-e (in.p |Rome|))};
   
\noindent
these constraints are ultimately cashed out as predications about episodes characterized
by the sentence being modified. (This is also true for the \ulf{past} and \ulf{pres}
tense operators.) Similarly any modifier of sort \ulf{.adv-f} is transparent
and implies the existence of a multi-episode (characterized by the sentence as a whole)
whose temporally disjoint parts each have the same characterization \citep{hwang1994ICTL}; e.g.,

  \ulf{regularly.adv-f}, \ulf{(adv-f ({at}.p (three.d (plur time.n))))};

\noindent
The earlier \textit{walk with Bob} 
example shows how in ULF the operator and operand can be inferred from the
constituent types.  Consider the types for \ulf{play.v} and \ulf{(adv-a (with.p (the.d dog.n)))}.  Since they have types $\prd_{V}$ and $\prd_{V} \rar \prd_{V}$, respectively, %
we can be certain that \ulf{(adv-a (with.p (the.d dog.n)))} is the operator while
\ulf{play.v} is the operand.

In practice, we're able to drop the \ulf{mod-a}, \ulf{mod-n}, and \ulf{nnp} 
type-shifters during annotation since we can post-process them with the appropriate 
type-shifter to make the composition valid. We assume in these cases that the prefixed
predicate is intended as the operator, which reflects a common pattern in English.  Thus, 
``burning hot melting pot'' would be hand annotated as

  \ulf{((burning.a hot.a) (melting.n pot.n))}

\noindent
which would be post-processed to

  \ulf{((mod-n ((mod-a burning.a) hot.a)) ((mod-n melting.n) pot.n))}

\noindent
While the prefixed predicate modification allows us to formally model non-intersective
modification, %
there are modification patterns in English that force an intersective interpretation, e.g., post-nominal modification and appositives, and we annotate them accordingly. 
\textit{``The buildings in the city''} is annotated

  \ulf{(the.d (n+preds (plur building.n)
                       (in.p (the.d city.n))))}

\noindent
which is equivalent (via the \ulf{n+preds} macro) to

  \ulf{(the.d (\lm x ((x (plur building.n)) and.cc
                            (x (in.p (the.d city.n))))))}.

\section{Annotating a ULF Corpus} 
\label{sec:annotation}

The syntactic relaxations in ULF and the annotation environment work hand-in-hand to enable quick and consistent annotations. ULF syntax relaxations are designed to: (1)~Preserve surface word order and (2)~Make the annotations match linguistic intuitions more closely.  As a result, annotating a sentence with its ULF interpretation boils down to marking the words with their semantic types, bracketing the sentence according to the operator-operand relations, then introducing macros and logical operators as necessary to make the ULF type-consistent.  The annotation environment is designed to assist in this process by improving the readability of long ULFs and catching mistakes that are easy to miss. The environment is shared across annotators with certainty marking so that more experienced annotators can correct and give feedback to trainees. This streamlines the training process and minimizes the mistakes entering into the corpus.
Here are the core annotator features.\footnote{The annotator can be accessed from the ULF project website and a screenshot of it is in Appendix~\ref{app:annotator}.}

\begin{enumerate}%
    \item \textbf{Syntax and bracket highlighting.} Highlights the cursor location and the closing bracket, unmatched brackets and quotes, operator keywords, and badly placed operators.
    \item \textbf{Sanity checker.} Alerts the annotator to invalid type compositions and suggests corrections for common mistakes.  
    \item \textbf{Certainty marking.} Annotators can mark whether they are certain of an annotation's correctness so that partial progress can be made while preserving the integrity of the corpus.
    \item \textbf{Sentence-specific comments.} Annotators can record their thoughts on partially complete annotations so that others can pick up where they left off.
\end{enumerate}

\noindent
The ULF type system makes it possible to build a robust sanity checker for the annotator.  The type system severely restricts the space of valid ULF formulas and usually when an annotator makes an error in annotation, it leads to a type inconsistency.  

\section{Experimental Results and Current Progress}
\label{sec:results-and-progress}

We ran a timing study and an interannotator agreement~(IA) study to quantify the efficacy of the presented annotation framework.  We timed 80~annotations of the Tatoeba dataset and found the average annotation speed to be 8~min/sent with 4~min/sent among the two experts and 11~min/sent among the three trainees that participated.  AMRs reportedly took on average 10~min/sent~\citep{hermjakob2013HLTdemo}.  In the IA study five annotators each annotated between 18 and 23 sentences from the same set of 23 sentences, marking their certainty of the annotations as they normally would.  The sentences were sampled from the four datasets listed in Table~\ref{table:ann-stats}.  The mean and standard deviation of sentence length were 15.3 words and 10.8 words, respectively.  

\begin{wrapfigure}[13]{r}{0.42\textwidth}
\vspace*{-0.1in}
\captionof{table}{{\small Current sentence annotation counts broken down by dataset and certainty.  DG and PG are the Discourse Graphbank and Project Gutenberg, respectively.  The \textit{Old} column annotations are from before we added the certainty feature.}}\label{table:ann-stats} 
\centering
\vspace*{-0.1in}
{\small
\tabcolsep=0.15cm
\begin{tabular}{ |l|l|l|l|l|l|l|l|l| } 
 \hline
  & \textit{Cert.} & \textit{Unc.} & \textit{Inc.} & \textit{Old} & \textit{All} \\ \hline 
 Tatoeba        & 533 & 66  & 24    & 396   & 1019  \\ 
 DG             & 102 & 37  & 4     & 0     & 143   \\
 UIUC QC        & 179 & 50  & 0     & 0     & 229   \\
 PG             & 113 & 59  & 17    & 0     & 189   \\ \hline
 Total          & \textbf{927} & 212 & 45    & 396   & 1580  \\
 \hline
\end{tabular}
}
\end{wrapfigure}

We computed a similarity score between two annotations using \textit{EL-smatch}~\citep{kim2016*SEM}, a generalization of \textit{smatch}~\citep{cai2013ACL} which handles non-atomic operators.
The document-level EL-smatch score between all annotated sentence pairs was 0.70.  When we restricted the analysis to just annotations that were marked \textit{certain}, the agreement rose to 0.78. The complete pairwise scores are shown in Table~\ref{table:ia-matrix}.  Notice that annotators 1, 2, and 3 had very high agreement with each other.  If we restrict the agreement to just those three annotators, the full and certain-subset scores are 0.79 and 0.88, respectively.  Out of all the annotations, less than a third were marked as uncertain or incomplete.  AMR annotations reportedly have annotator vs consensus IA of 0.83 for newswire and 0.79 for web text~\citep{tisalos2015slides}.

\begin{wrapfigure}[9]{r}{0.46\textwidth}
\vspace*{-0.1in}
\captionof{table}{{\small Pairwise IA scores, where the left score is over all annotations and the right score is only over annotations marked as certain.}}\label{table:ia-matrix} 
\centering
\vspace*{-0.1in}
{\small
\tabcolsep=0.15cm
\begin{tabular}{ |l|l|l|l|l| } 
 \hline
    & 2         & 3         & 4         & 5         \\ \hline 
 1  & 0.80/0.88 & 0.79/0.89 & 0.69/0.77 & 0.63/0.75 \\ \hline
 2  & -         & 0.77/0.86 & 0.72/0.77 & 0.62/0.75 \\ \hline
 3  & -         & -         & 0.69/0.75 & 0.63/0.73 \\ \hline
 4  & -         & -         & -         & 0.62/0.71 \\ \hline
\end{tabular}
}
\end{wrapfigure}

This study also demonstrates that the certainty marking indeed reflects the quality of the annotation, thus performing the role we intended.  Also, based on the high agreement between annotators 1, 2, and 3, we can conclude that consistent ULF annotations across multiple annotators is possible.  However, the lower scores of annotators 4 and 5, even in annotations marked as certain, indicates room for improvement in the annotation guidelines and training of some annotators.

We have so far collected 927 certain annotations and have 1,580 in total. The full annotation breakdown is in Table~\ref{table:ann-stats}.  
We started with the English portion of the Tatoeba dataset~(\url{https://tatoeba.org/eng/}), a crowd-sourced translation dataset.  
This source tends to have shorter sentences, but they are more varied in topic and form.
We then added text from Project Gutenberg~(\url{http://gutenberg.org}), the UIUC Question Classification dataset~\citep{li2002COLING}, and the Discourse Graphbank~\citep{wolf2005thesis}.  
Preliminary parsing experiments on a small dataset (900 sentences) show promising results 
and we expect to be able to build
an accurate parser with a moderately-sized dataset and representation-specific engineering~\citep{kim2019FLAIRS}.

\section{Related Work}
\label{sec:related-work}

A notable development in general representations of semantic content has been the design of AMR~\citep{banarescu2013LAW} followed by numerous research studies on generating AMR from English and on using it for downstream tasks. 
AMR is intended as a kind of intuitive normal form for the relational context of English sentences in order to assist in machine translation.
Given this goal, AMR deliberately neglected issues such as articles, tense, the distinction between real and hypothetical entities, and non-intersective modification. In the context of inference, this risks making false conclusions such as that a \textit{``big ant''} is bigger than a \textit{``small elephant''}.

Still, this development was an inspiration to us in terms of both the quest for broad 
coverage and methods of learning and evaluating semantic parsers.
There has also been much activity in developing semantic
parsers that derive logical representations, raising the possibility
of making inferences with those representations~\citep{artzi2015EMNLP,artzi2013TACL,howard2014ICRA,kate2006COLACL,konstas2017ACL,kwiatkowski2011EMNLP,liang2011ACL,poon2013ACL,popescu2004COLING,tellex2011AAAI}. 
The techniques and formalisms employed are interesting (e.g., learning
of CCG grammars that generate $\lambda$-calculus expressions), but 
the targeted tasks have generally been question-answering in
domains consisting of numerous monadic and dyadic ground facts
(``triples"), or simple robotic or human action descriptions.\footnote{%
  For example, \cite{ross2018EMNLP} develop a CCG-based semantic parser for action annotations in videos, 
  representing sentences in an approximate way---neglecting determiners and treating all entity references as variables.}

Noteworthy examples of formal logic-based approaches, not targeting
specific applications are Bos'~(\citeyear{bos2008STEP}) and Draiccio et al.'s~(\citeyear{draicchio2013ESWC}),
whose hand-built semantic parsers respectively generate FOL formulas and
OWL-DL expressions. But these representations preclude generalized
quantifiers, modification, reification, attitudes, etc.
We are not aware of any work on inference generation of the type ULFs targets, based on these projects. A couple yet-unmentioned but notable semantic annotation projects are the Groningen Meaning Bank~\citep{bos2017GMB}, with discourse representation structure~(DRS) annotations~\citep{kamp1981FMSL}
and the Redwoods treebank~\citep{flickinger2012ITLT,oepen2002COLING} with Minimal Recursion Semantics~(MRS)~\citep{copestake2005RLC} annotations. DRSs have the same representational limitations as Bos'~(\citeyear{bos2008STEP}) system. MRS is descriptively powerful and linguistically motivated, with significant resources including a hand-built grammar, multiple parsers, and a large annotated dataset~\citep{bub1997ICASSP,callmeier2001thesis}. Given that MRS is an object-language agnostic, meta-level semantic representation, an inference system cannot be built directly for MRS based on model-theoretic notions of interpretation, truth, satisfaction, and entailment. However, the lack of an object-language in MRS leaves open the possibility of forming a correspondence between MRS and ULF that fully respects both formalisms.
Finally, the use of unscoped LFs in a rule-to-rule framework was first introduced by~\cite{schubert1982CL} and a similar approach to scope ambiguity was taken by the Core Language Engine~\citep{alshawi1989ACL}.

\section{Conclusion \& Future Work}
\label{sec:conclusion}

ULF, the underspecified initial representation for EL described in this document, captures a subset of the semantic information of EL that allows it to be annotated reliably, participate in the complete resolution to EL, and form the basis for structural inferences that are important for language understanding tasks.  We will continue this work by expanding the corpus of ULF annotations and training a statistical parser over that corpus. Automatic ULF parses could then be used as the backbone for a complete EL parser or as the core representation for NLP tasks that require sentence-level formal semantic information or structural inferences.

\section{Acknowledgements}

We would like to thank Burkay Donderici, Benjamin Kane, Lane Lawley, Tianyi Ma, Graeme McGuire, Muskaan Mendriatta, Akihiro Minami, Georgiy Platonov, Sophie Sackstein, and Siddharth Vashishta for raising thoughtful questions in the development of this work. We are grateful to the anonymous reviewers for their helpful feedback. This work was supported by DARPA CwC subcontract W911NF-15-1-0542.

\bibliographystyle{chicago}
\bibliography{main}

\appendix

\section{Quantifier Semantics}
\label{app:quantifier-semantics}

Noun phrases can occur in any position 
here an individual variable or constant can occur, and in post-processing are replaced by bound variables. Therefore the \textit{positional} types of noun phrases are individuals, \dom. Therefore, we can treat determiners such as \ulf{every.d} in ULF as if they were of type $(\prd \rar \dom$, i.e. a function from a predicate to an individual.  For example consider the ULF formula \ulf{((every.d dog.n) (pres run.v))}.  \ulf{(every.d dog.n)} seems to be able to occur in any place that \ulf{|John|} and \ulf{they.pro} can occur.  
    
\begin{tabular}{ r l }
    \ulf{((every.d dog.n) (pres run.v))},   & \ulf{(i.pro ((pres like.v) (every.d dog.n)))}, \\

    \ulf{(|John| (pres run.v))},           & \ulf{(i.pro ((pres like.v) |John|))}, \\
    
    \ulf{(they.pro (pres run.v))};        & \ulf{(i.pro ((pres like.v) they.pro))};
\end{tabular}

\noindent
Semantically we consider \ulf{they.pro} and \ulf{them.pro} to be the same, as they only differ in syntactic position.  Then since \ulf{dog.n} (and any other argument of a determiner) is a monadic predicate, we can infer that the \textit{positional} type of determiners is $\prd \rar \tru$.  This will be transformed after scoping into a formula of the form $(\delta v: \phi \; \psi)$, where $\delta$ is the determiner, and $\phi$ and $\psi$ correspond to the formulas resulting from substituting the scoped variable into the restrictor and matrix predicates, respectively.  These formulas are interpreted in EL via satisfaction conditions over the quantified variable and two formulas (a restrictor formula and the nuclear scope), e.g., for an sentence such as \textit{``Most car crashes are due to driver error"},%

$(\text{most } v: \phi \; \psi)^{\model \varas} = 1$ iff\\ \hspace*{3em} for most $d \in \dom$ 
   for which $\phi^{\model \varas_{v:d}} = 1$, $\psi^{\model \varas_{v:d}} = 1$

\noindent
where \model is the model, \varas is the variable assignment function, and $\varas_{v:d}$ is the same as $\varas$ except that its value for variable $v$ is $d$. %
When this formula is evaluated with respect to an episode, it corresponds to a formula of the form

$[(\text{several } v: \phi \; \psi) \; {**} \; \eta]$,

\noindent
where `$**$' is the operator relating a sentence to the episode it \textit{characterizes} (describes as a whole), which is discussed in Section~\ref{sec:el}.  $(\delta v: \phi \; \psi)$ can equivalently be rewritten as $(\delta \; (\lm v \; \phi) \; (\lm v \; \psi))$ and we can define $\delta$ as a second-order intensional predicate of type $\prd \rar \prd \rar \sit \rar \tru$ similar to the approach used in generalized quantifier theory~\citep{barwise1981generalized}.  

\section{Episodic Operators}
\label{app:episodic-ops}

`**', `*', and `@' are \textit{episodic} operators, which relate formulas to episode variables in Episodic Logic.  They do not appear in ULFs since ULFs do not have explicit episode variables.  However, these operators are foundational to Episodic Logic semantics in handling event structure and intensional semantics.  All formulas in EL must be evaluated with respect to one of these operators to obtain a truth value since sentence intensions in EL have the type $\sit \rar \tru$.

\begin{itemize}[leftmargin=*]
    \item `**' - the \textit{characterizing} operator
    
    `**' relates an episode variable to a formula that \textit{characterizes} it. In other word, the formula describes the episode as a whole, or the nature of the episode, rather than a tangential part or a temporal segment of it.  This, however, does not mean that the characterizing formula must describe \textit{every} detail of the episode.  It can in fact be quite abstract.  For instance, \textit{``John had a car accident''} and \textit{``John hit some black ice and his car skidded into a tree''} might characterize the same event.  As such, for most news stories the headline and the first sentence of the article are likely to both characterize the same event even though the headline is much shorter.  Formally, 
    $$[\phi \; {**} \; \eta]^{\model \varas} = 1 \text{ iff } \phi^{\model \varas}(\eta^{\model \varas}) = 1;$$
    $$[(\text{not }\phi) \; {**} \; \eta]^{\model \varas} = 1 \text{ iff } \phi^{\model \varas}(\eta^{\model \varas}) = 0.$$
    
    The semantic type of $\phi$ is $\sit \rar \tru$ (a sentence intension) and the semantic type of $\eta$ is $\sit$, a situation.  Therefore, $\eta$ characterizes $\phi$ just in the case that the interpretation of $\phi$ with respect to the model $\model$ and variable assignment function $\varas$ evaluated over the interpretation of $\eta$ with respect to $\model$ and $\varas$ is true.
    
    \item `*' - the \textit{truth} operator
    
    `*' relates an episode variable to a formula that is \textit{true} in that episode.  This is a weaker operator than `**' in that a formula that is `*'-related can be a just a segment or an incidental aspect of the episode to be true.  Therefore, $[\phi \; {**} \; \eta]$ entails $[\phi \; * \; \eta]$, but not the other way.  Therefore, \textit{``There was black ice on the road''} and \textit{``John was driving''} could both be `*'-related to the episode characterized by the example given in for the `**' operator. Formally,
    $$[\phi \; * \; \eta]^{\model \varas} = 1 \text{ iff there is an episode }s \sqsubseteq \eta^{\model \varas}\text{ such that } \phi^{\model \varas}(s) = 1.$$
    
    Where $\sqsubseteq$ is an episode part-of relation.  It's formal definition is given by \cite{hwang1993STA}.  Intuitively we can think of $s \sqsubseteq \eta$ to mean that $s$ is a subepisode of $\eta$.
    
    \item `@' - the \textit{concurrent} operator
    
    `@' relates an episode variable to a formula characterizes another episode that runs concurrent with it.  So this operator can be rewritten in the following way.  $[\phi \; @ \; \eta]$ entails and is entailed by $(\text{some }e: [e \text{ same-time } \eta] \; [\phi \; {**} \; e])$.  Formally, $@$ us defined as 
    $$[\phi \; @ \; \eta]^{\model \varas} = 1 \text{ iff there is an episode }s \in \sit\text{ with \textit{time}}(s) = \text{\textit{time}}(\eta^{\model \varas})\text{ such that }\phi^{\model \varas}(s) = 1.$$
    
\end{itemize}

\section{More About Macros}
\label{app:macro}

ULF macros are different syntactic rewriting operators to reduce the annotator burden of encoding complex, but regular, semantic structures or avoid unnecessary word reordering.  Table~\ref{table:macros} lists the definitions and simple examples of the basic ULF macros.  The \ulf{sub} macro is the \textit{substitution} macro which performs a simple substitution of its first argument into the position of \ulf{*h} within the second argument.  This is used for topicalization, such as \textit{``Swiftly, the fox ran away''}, which topicalizes \textit{``Swiftly''} from the sentence \textit{``The fox swiftly ran away''}.  The \ulf{rep} macro is the \textit{replace} operator and the exact same as \ulf{sub} with the arguments swapped and using \ulf{*p} instead of \ulf{*h} as the placeholder variable.  This is used for rightward-displaced clauses, such as, \textit{``A man answered the door with a white beard''}, in which \textit{with a white beard} is really displaced from the expected post-nominal position, i.e \textit{``A man with a white beard ...''}.

\begin{figure}[ht]
\captionof{table}{\label{table:macros} List of basic rewriting macros in ULF.  $=_m$ is the macro defining operator.}
\centering
{\small
\tabcolsep=0.11cm
\begin{tabular}{ |l|l|l| } 
 \hline
 \textit{Name} & \textit{Definitions} & \textit{Example} \\ \hline 
 \ulf{sub} & \ulf{(sub C S[*h]) $=_m$ S[*h$\leftarrow$C]} & \ulf{(sub A (B *h))} $=_m$ \ulf{(B A)}\\ \hline
            
 \ulf{rep} & \ulf{(rep S[*p] C) $=_m$ S[*p$\leftarrow$C]} & \ulf{(rep (A *p) B)} $=_m$ \ulf{(A B)}\\ \hline
 \multirow{2}{*}{\ulf{n+preds}} & \ulf{(n+preds N P$_1$ ... P$_n$)} $=_m$ & 
                  \ulf{(n+preds dog.n red.a)} $=_m$\\ 
    & \ulf{\numCharTab{2}(\lm x ((x N) and.cc (x P$_1$) ... (x P$_n$)))} & 
    \ulf{\numCharTab{2}(\lm x ((x dog.n) and.cc (x red.a)))}\\ \hline                  
 \multirow{3}{*}{\ulf{np+preds}} & \ulf{(np+preds NP P$_1$ ... P$_n$)} $=_m$ & 
                  \ulf{(np+preds he.pro red.a)} $=_m$\\
    & \ulf{\numCharTab{2}(the.d (\lm x ((x = NP) and.cc}  & 
    \ulf{\numCharTab{2}(the.d (\lm x ((x = he.pro) and.cc}\\ 
    
    & \ulf{\numCharTab{14}(x P$_1$) ... (x P$_n$))))} &
    \ulf{\numCharTab{14}(x red.a))))} \\
    \hline              
 \multirow{2}{*}{\ulf{'s}}  & \multirow{2}{*}{\ulf{((NP 's) N)} $=_m$ \ulf{(the.d ((poss-by NP) N))}} &
                  \ulf{((|John| 's) dog.n)} $=_m$ \\
                  & & \ulf{\numCharTab{2}(the.d ((poss-by |John|) dog.n))} \\
 \hline
\end{tabular}
}
\end{figure}

Next, \ulf{n+preds} and \ulf{np+preds} are macros for handling post-nominal modification.  \ulf{n+preds} modifies a noun and returns a noun, whereas \ulf{np+preds} modifies an entity and returns a modified entity. Intuitively, \ulf{np+preds} handles non-restrictive modifiers, whereas \ulf{n+preds} handles restrictive modifiers.  This makes sense since the modifying predicates in \ulf{n+preds} are added before the determiner, thus introduced into the restrictor of the quantification.

\ulf{'s} is a macro for handling possession using an appended marker to the possessor just as is done in English (e.g. ``John's dog'').  Formally, this maps to a pre-modifying possession relation.  So \textit{``John's dog''} is hand-annotated as \ulf{((|John| 's) dog.n)}, which expands out to \ulf{(the.d ((poss-by |John|) dog.n))}.   \ulf{poss-by} is a binary predicate relating two entities, semantic type $\dom \rar (\dom \rar (\sit \rar \tru))$.  so \ulf{(poss-by |John|)} resolves to semantic type of a predicate, \prd.  Notice that this is a predicate-noun pair so as discussed in Section~\ref{sec:ulf-details} the \ulf{mod-n} type-shifter is automatically introduced, resulting in \ulf{(the.d ((mod-n (poss-by |John|)) dog.n))}. 

\pagebreak

\section{Additional Annotator Info}
\label{app:annotator}

\begin{wrapfigure}[25]{r}{0.5\textwidth}
\begin{minipage}{\linewidth}
\includegraphics[width=\linewidth]{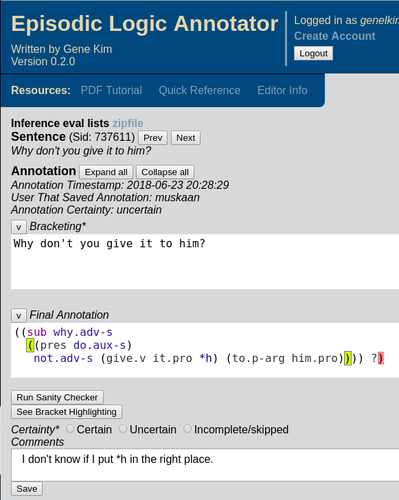}
\caption{Current ULF annotator state with example annotation process.}
\label{fig:ulf-annotator}
\end{minipage}
\end{wrapfigure}

Here we reiterate the annotator features as described in Section~\ref{sec:annotation} with reference to an image of it in Figure~\ref{fig:ulf-annotator}.

\begin{enumerate}[leftmargin=*,noitemsep,topsep=0pt]
    \item \textbf{Syntax and bracket highlighting.} Highlights the cursor location and the closing bracket, unmatched brackets and quotes, operator keywords, and badly placed operators.  The ``Final Annotation'' window in Figure~\ref{fig:ulf-annotator} shows the cursor matching bracket in yellow-green highlighting, an unmatched bracket in red, the \ulf{sub} macro in purple, and sentence-level operators in blue. 
    \item \textbf{Sanity checker.} Alerts the annotator to invalid type compositions and suggests corrections for common mistakes.  
    \item \textbf{Certainty marking.} Annotators can mark whether they are certain of an annotation's correctness so that partial progress can be made while preserving the integrity of the corpus.  The bottom of Figure~\ref{fig:ulf-annotator} shows radio buttons for selecting the certainty of the annotation.
    \item \textbf{Sentence-specific comments.} Annotators can record their thoughts on partially complete annotations so that others can pick up where they left off.  The bottom-most window in view in Figure~\ref{fig:ulf-annotator} is the sentence-specific comment window.  These comments are viewable by all annotators when accessing this sentence.
\end{enumerate}

\section{Additional Grounding Examples}

Here are a couple of additional sections that ground the high-level ULF background in concrete examples.  

\subsection{More Resources on Predicate Modifiers}

A type of modification not covered in the main document is entity-predicate modification.  The type shifter from an individual to a nominal predicate modifier is named \ulf{nnp} and has semantic type, $\dom \rar (\prd_{N} \rar \prd_{N})$.  It is for indicating premodification of a common noun by a proper noun; e.g.,

   \ulf{((nnp |Seattle|) skyline.n)}.

\noindent
All of the operators discussed in Section~\ref{sec:ulf-details} and here are listed alongside a ULF example, and its semantic type in Table~\ref{table:pred-mod-formers}.

\begin{figure}[ht]
\captionof{table}{\label{table:pred-mod-formers}Predicate and sentence 
modifier forming operators in ULF along with examples and their semantic types.}
\centering
\begin{tabular}{ |l|l|l| } 
 \hline
 \textit{Name} & \textit{Example} & \textit{Semantic Type} \\ \hline 
 \ulf{mod-a}    & \ulf{((\underline{mod-a} worldly.a) wise.a)} & 
                  $\prd \rar (\prd_{ADJ} \rar \prd_{ADJ})$ \\ 
 \ulf{mod-n}    & \ulf{((\underline{mod-n} (very.mod-n happy.a)) dog.n)} & 
                  $\prd \rar (\prd_{N} \rar \prd_{N})$ \\
 \ulf{adv-a}    & \ulf{(play.v (\underline{adv-a} (with.p (a.d dog.n))))} & 
                  $\prd \rar (\prd_{V} \rar \prd_{V})$ \\
 \ulf{nnp}      & \ulf{((\underline{nnp} |Seattle|) skyline.n)} & 
                  $\dom \rar (\prd_{N} \rar \prd_{N})$ \\ \hline
 \ulf{adv-s}    & \ulf{(show\_up.v (\underline{adv-s} (to.p (my.d surprise.n))))} &
                  $(\sit \rar \tru) \rar (\sit \rar \tru)$ \\
 \ulf{adv-e}    & \ulf{(eat.v (\underline{adv-e} (at.p (a.d cafe.n))))} &
                  $(\sit \rar \tru) \rar (\sit \rar \tru)$ \\
 \ulf{adv-f}    & \ulf{(run.v (\underline{adv-f} (very.mod-a often.a)))} &
                  $(\sit \rar \tru) \rar (\sit \rar \tru)$ \\
 \hline
\end{tabular}
\end{figure}

\noindent
Ultimately  in EL, \ulf{adv-a}, \ulf{adv-e}, and \ulf{adv-f} will be reconstrued as predications over actions and events via meaning postulate inferences.  
Agent-episode pairs that intuitively represent actions, experiences, or attributes are distinct from events.  For example, \textit{``He fell \underline{painfully}"} refers to a painful experience rather than to a painful event and \textit{``He excels \underline{intellectually}"} refers an intellectual attribute rather than to an intellectual event or situation.  \ulf{.adv-a} type modifiers constrain both the agent and the episode in the pair.  No sharp or exhaustive classification of such pairs into actions, experiences, and attributes is presupposed by this -- the point is just to make available the subject of sentences in working out entailments of VP-modification.  Since actions are formed by pairing an agent with an event variable, they are not explicitly represented in ULF.  The meaning postulate inferences on \ulf{.adv-a} type modifiers would infer from \ulf{(he.pro (play.v (adv-a (with.p (a.d dog.n)))))} the following deindexed ULF \ulf{[[[he.pro play.v] ** E1.sk] and.cc [(pair he.pro E1.sk) (with.p (a.d dog.n))]]}.  The meaning postulate inference of \ulf{.adv-e} type modifiers to predications over events is also straightforward.  The ULF formula \ulf{(she.pro (eat.v (adv-e (at.p (a.d cafe.n)))))} leads to the deindexed, inferenced formula \ulf{[[[she.pro eat.v] ** E1.sk] and.cc [(pair she.pro E1.sk) (at.p (a.d cafe.n))]]}.  

\subsection{Topicalization \& Relative Clauses in ULF}
The \ulf{sub} macro was introduced to reduce the amount of lexical reordering
by annotators when annotating sentences 
with syntactic movement such as topicalization. 
\ulf{sub} takes two constituents, the 
second of which must contain the symbol \ulf{*h}.  When the operator is evaluated 
the first argument is inserted into the position of \ulf{*h} in the second argument.
\textit{``Swiftly, the fox ran away''} for example would be annotated as (in raw ULF form)

  \ulf{(sub swiftly.adv-a ((the.d fox.n) ((past run.v) away.adv-a *h)))}

\noindent
and when the \ulf{sub} macro is evaluated, becomes

  \ulf{((the.d fox.n) ((past run.v) away.adv-a swiftly.adv-a))}.

\noindent
For relative clauses we introduce one extra post-processed element which is the 
relativizer, annotated with a \ulf{.rel} extension.  \textit{``The coffee that you drank''} 
is annotated in raw ULF with macros as

  \ulf{(the.d (n+preds coffee.n (sub that.rel (you.pro ((past drink.v) *h)))))}

\noindent
During post-processing, the embedded sentence in which the \ulf{.rel} variable
lies is \lm-abstracted and the lambda variable replaces the \ulf{.rel}
variable.  Post-processing \ulf{that.rel} leads to

  \ulf{(the.d (n+preds coffee.n (\lm x (sub x (you.pro ((past drink.v) *h))))))}

\noindent
Now if we evaluate both \ulf{n+preds} and \ulf{sub}, and perform one lambda reduction we get

  \ulf{(the.d (\lm y ((y coffee.n) and.cc (you.pro ((past drink.v) y)))))}

\noindent
which is exactly the meaning that is expected that is expected from the relative clause.  That is, \textit{``The coffee that you drank''} is a coffee~(\ulf{(y coffee.n)}) and is something that you drank~(\ulf{(you.pro ((past drink.v) y))}).

\end{document}

%% file: semantic-interpretation.tex
\pgfdeclarelayer{background}
\pgfdeclarelayer{foreground}
\pgfsetlayers{background,main,foreground}

\newcommand{\dopacity}{10}
\newcommand{\hopacity}{20}

\newcommand{\smulf}[1]{\ulf{\fontsize{9.5}{9.5}\selectfont{#1}}}

\tikzstyle{defarrow}=[>=latex, decoration={markings,mark=at position 1 with {\arrow[thick]{>}}},
    postaction={decorate}]
\tikzstyle{strarrow}=[defarrow, thick, draw]
\tikzstyle{infarrow}=[defarrow, dashed, draw=blue]

\tikzstyle{stage}=[text centered, text width=7em, fill=none, 
    draw=none, minimum height=3em, 
    rounded corners] %
\tikzstyle{display}=[draw=none, text width=21.5em, minimum height=1em,inner sep=0.5em]
\tikzstyle{ministage}=[draw=black!50, dashed, fill=orange!\dopacity, inner sep=0.5em, rounded corners, minimum height=1em, text width=7.8em]
\tikzstyle{key}=[text width=8.5em, minimum height=1em, rounded corners, inner sep=0.5em, draw=black]

\def\blockdist{4.4}
\def\edgedist{2.5}

\begin{tikzpicture}
    \node (eng) [stage]  {\textbf{English}};
    \path (eng.east)+(\blockdist,0) node (engex) [display] {\textit{She wants to eat the cake.}};
    \begin{pgfonlayer}{background}
        \path (eng.west |- engex.north) node (a) {};
        \path (engex.east |- engex.south) node (b) {};
        \path[fill=yellow!\dopacity,rounded corners, draw=black!50, dashed]
            (a) rectangle (b);           
        \path (eng.east |- engex.south) node (c) {};
        \path[fill=blue!\dopacity,rounded corners, draw=black!50, draw=none]
            (a) rectangle (c);           
        \path[fill=blue!\dopacity, draw=none]
            (eng.west |- engex.north) + (1.0,0) rectangle (c);           
    \end{pgfonlayer}
    \path[fill=none, rounded corners, draw=black!50, dashed]
        (a) rectangle (b);           
   
    \path (eng.south)+(0,-0.9) node (ulf) [stage] 
    {\textbf{ULF \\
    (Unscoped)}};
    \path (ulf.east)+(\blockdist,0) node (ulfex) [display] 
    {\smulf{(she.pro ((pres want.v)\\
    \hspace*{4.9em}(to (eat.v (the.d cake.n)))))}};
    \begin{pgfonlayer}{background}
        \path (ulf.west |- ulfex.north)+(0,0) node (a) {};
        \path (ulfex.east |- ulfex.south) node (b) {};
        \path[fill=yellow!20,rounded corners, draw=black, thick, drop shadow]
            (a) rectangle (b);           
        \path (ulf.east |- ulfex.south) node (c) {};
        \path[fill=blue!20,rounded corners, draw=black!50, draw=none]
            (a) rectangle (c);           
        \path[fill=blue!20, draw=none]
            (ulf.west |- ulfex.north) + (1.0,0) rectangle (c);           
    \end{pgfonlayer}
    \path[fill=none, rounded corners, draw=black, thick]
        (a) rectangle (b);           
    
    \path (ulf.south)+(0,-1.2) node (slf) [stage] {
    \textbf{SLF \\
    (Scoped)}};
    \path (slf.east)+(\blockdist,0) node (slfex) [display] 
    {\smulf{(\colorbox{yellow}{pres} (\colorbox{yellow}{the.d x (x cake.n)}\\
    \hspace*{4.9em}(she.pro (want.v (to (eat.v \colorbox{yellow}{x}))))))}};
    \begin{pgfonlayer}{background}
        \path (slf.west |- slfex.north)+(0,0) node (a) {};
        \path (slfex.east |- slfex.south) node (b) {};
        \path[fill=yellow!\dopacity,rounded corners, draw=black!50, dashed]
            (a) rectangle (b);           
        \path (slf.east |- slfex.south) node (c) {};
        \path[fill=blue!\dopacity,rounded corners, draw=black!50, draw=none]
            (a) rectangle (c);           
        \path[fill=blue!\dopacity, draw=none]
            (slf.west |- slfex.north) + (1.0,0) rectangle (c);           
    \end{pgfonlayer}
    \path[fill=none, rounded corners, draw=black!50, dashed]
        (a) rectangle (b);           
   
    \path (slf.south)+(0,-1.6) node (clf) [stage] 
    {\textbf{CLF\\
     (Contextual)}};
    \path (clf.east)+(\blockdist,0) node (clfex) [display] 
    {\smulf{(\colorbox{yellow}{|E|.sk at-about.p |Now17|})},\\
    \smulf{((the.d x (x cake.n) \\
    \hspace*{2em}(she.pro (want.v (to (eat.v x))))) \colorbox{yellow}{** |E|.sk})}};
    \begin{pgfonlayer}{background}
        \path (clf.west |- clfex.north)+(0,0) node (a) {};
        \path (clfex.east |- clfex.south) node (b) {};
        \path[fill=yellow!\dopacity,rounded corners, draw=black!50, dashed]
            (a) rectangle (b);           
        \path (clf.east |- clfex.south) node (c) {};
        \path[fill=blue!\dopacity,rounded corners, draw=black!50, draw=none]
            (a) rectangle (c);           
        \path[fill=blue!\dopacity, draw=none]
            (clf.west |- clfex.north) + (1.0,0) rectangle (c);           
    \end{pgfonlayer}
    \path[fill=none, rounded corners, draw=black!50, dashed]
          (a) rectangle (b);           
   
    \path (slf.west)+(-2.5,0) node (dix) [ministage] 
    {\smulf{x} $\rightarrow$ \smulf{|Cake3|}, \\
     \smulf{she.pro} $\rightarrow$ \smulf{|Chell|}};
    \node [left, xshift=0.5em, yshift=0.5em] at (dix.north) {\textit{Anaphora}};
  
    \path (clf.west)+(-2.5,0) node (wsd) [ministage] 
    {\smulf{want.v} $\rightarrow$ \smulf{want1.v},\\
     \smulf{eat.v} $\rightarrow$ \smulf{eat1.v},\\
     \smulf{cake.n} $\rightarrow$ \smulf{cake1.n}};
  \node [left, xshift=-1.4em, yshift=0.5em] at (wsd.north) {\textit{WSD}};
  
    \path (clf.south)+(-1.8,-1.4) node (elf) [stage] 
    {\textbf{ELF\\
    (Episodic)}};
    \path (elf.east)+(\blockdist,0) node (elfex) [display] 
    {\smulf{(|E|.sk at-about.p |Now17|)},\\
    \smulf{((\colorbox{yellow}{|Chell|} (\colorbox{yellow}{want1}.v (to (\colorbox{yellow}{eat1}.v \colorbox{yellow}{|Cake3|}))))\\
    \hspace*{1em}** |E|.sk)}};
    \begin{pgfonlayer}{background}
        \path (elf.west |- elfex.north)+(0,0) node (a) {};
        \path (elfex.east |- elfex.south) node (b) {};
        \path[fill=yellow!\dopacity,rounded corners, draw=black!50, dashed]
            (a) rectangle (b);           
        \path (elf.east |- elfex.south) node (c) {};
        \path[fill=blue!\dopacity,rounded corners, draw=black!50, draw=none]
            (a) rectangle (c);           
        \path[fill=blue!\dopacity, draw=none]
            (elf.west |- elfex.north) + (1.0,0) rectangle (c);           
    \end{pgfonlayer}
    \path[fill=none, rounded corners, draw=black!50, dashed]
       (a) rectangle (b);           
   
    \path (dix.north)+(0.2,2.0) node (key) [key] 
        {{\small structure flow \\
                 information flow}};
    \node [left, xshift=-2em, yshift=0.5em] at (key.north) {\textbf{Key}};
    \path [->, strarrow] ($(key.north) + (1.1,-0.35)$) to ($(key.north) + (1.6,-0.35)$);
    \path [->, infarrow] ($(key.north) + (1.1,-0.8)$) to ($(key.north) + (1.6,-0.8)$);
    
    \newdimen\labx
    \newdimen\exx
    \newdimen\engbot
    \newdimen\ulftop
    \newdimen\ulfbot
    \newdimen\slftop
    \newdimen\slfbot
    \newdimen\clftop
    \newdimen\clfbot
    \newdimen\elftop
    \newdimen\elfx
    \newdimen\temp
    
    \newdimen\engx
    \newdimen\engy
    \path (eng.south); \pgfgetlastxy{\labx}{\engbot}
    \path (engex.south); \pgfgetlastxy{\exx}{\engbot}
    \path (ulfex.north); \pgfgetlastxy{\exx}{\ulftop}
    \path (ulfex.south); \pgfgetlastxy{\exx}{\ulfbot}
    \path (slfex.north); \pgfgetlastxy{\exx}{\slftop}
    \path (slfex.south); \pgfgetlastxy{\exx}{\slfbot}
    \path (clfex.north); \pgfgetlastxy{\exx}{\clftop}
    \path (clfex.south); \pgfgetlastxy{\exx}{\clfbot}
    \path (elfex.north); \pgfgetlastxy{\temp}{\elftop}
    \path (elf.north); \pgfgetlastxy{\elfx}{\temp}
    
    \newcommand{\forwshift}{0mm}
    \newcommand{\backshift}{-2mm}
    
	\path [->, transform canvas={xshift=\forwshift}, strarrow] 
	    (\labx, \engbot) -- node [right] 
	    {\textit{Parsing}} 
        (\labx, \ulftop);
	\path [->, transform canvas={xshift=\forwshift}, strarrow] 
	    (\labx, \ulfbot) -- node [right] 
	    {\textit{Scoping}} 
        (\labx, \slftop);
    \path [->, transform canvas={xshift=\backshift}, infarrow] 
        (\labx, \slftop) -- (\labx, \ulfbot) ;
	\path [->, transform canvas={xshift=\forwshift}, strarrow] 
	    (\labx, \slfbot) -- node [right] 
	    {\textit{Deindexing} and \textit{Canonicalization}} 
        (\labx, \clftop);
    \path [->, transform canvas={xshift=\backshift}, infarrow] 
        (\labx, \clftop) -- (\labx, \slfbot);
	\path [->, strarrow] 
        (\labx, \clfbot) -- node [right, yshift=2em]
        {}
        (\elfx, \elftop);

    \newcommand{\inshift}{0,4mm}
    \newcommand{\outshift}{0,-4mm}

    \path [->, infarrow]  (eng.west) -- ($(dix.east) + (\inshift)$);
    \path [->, strarrow]  (ulf.west) -- ($(dix.east) + (\inshift)$);
    \path [->, infarrow] ($(slf.west) + (\inshift)$) -- ($(dix.east) + (\inshift)$);
    \path [->, infarrow] ($(dix.east) + (\outshift)$) -- ($(slf.west) + (\outshift)$);
    \path [->, infarrow] ($(clf.west) + (\inshift)$) -- ($(dix.east) + (\inshift)$);
    \path [->, infarrow] ($(dix.east) + (\outshift)$) -- ($(clf.west) + (\outshift)$);
    \path [->, strarrow] ($(dix.east) + (\outshift)$) -- (\elfx, \elftop);
    
    \path [->, infarrow]  (eng.west) -- ($(wsd.east) + (\inshift)$);
    \path [->, strarrow]  (ulf.west) -- ($(wsd.east) + (\inshift)$);
    \path [->, infarrow] ($(slf.west) + (\inshift)$) -- ($(wsd.east) + (\inshift)$);
    \path [->, infarrow] ($(wsd.east) + (\outshift)$) -- ($(slf.west) + (\outshift)$);
    \path [->, infarrow] ($(clf.west) + (\inshift)$) -- ($(wsd.east) + (\inshift)$);
    \path [->, infarrow] ($(wsd.east) + (\outshift)$) -- ($(clf.west) + (\outshift)$);
    \path [->, strarrow] ($(wsd.east) + (\outshift)$) -- (\elfx,\elftop);

\end{tikzpicture}